\documentclass[10pt,twocolumn,letterpaper]{article}

\usepackage{iccv}
\usepackage{times}
\usepackage{epsfig}
\usepackage{graphicx}
\usepackage{amsmath}
\usepackage{amssymb}
\usepackage{tabulary}
\usepackage{multirow}
\usepackage{silence}
\usepackage{caption}

\usepackage[pagebackref=true,breaklinks=true,letterpaper=true,colorlinks,bookmarks=false]{hyperref}

\newcolumntype{x}[1]{>{\centering\arraybackslash}p{#1pt}}

\newlength\savewidth\newcommand\shline{\noalign{\global\savewidth\arrayrulewidth
  \global\arrayrulewidth 1pt}\hline\noalign{\global\arrayrulewidth\savewidth}}
\newcommand{\tablestyle}[2]{\setlength{\tabcolsep}{#1}\renewcommand{\arraystretch}{#2}\centering\footnotesize}

\iccvfinalcopy 

\def\iccvPaperID{2681} 

\ificcvfinal\fi

\begin{document}

\title{Animatable Neural Radiance Fields for Modeling Dynamic Human Bodies}

\author{
Sida Peng$^1$$^*$
\quad
Junting Dong$^1$$^*$
\quad
Qianqian Wang$^2$
\quad
Shangzhan Zhang$^1$ \\[1.5mm]
\quad
Qing Shuai$^1$
\quad
Xiaowei Zhou$^1$
\quad
Hujun Bao$^{1\dagger}$
\\[1.5mm]
$^1$Zhejiang University
\quad
$^2$Cornell University
}

\twocolumn[\maketitle\vspace{-3em}
\begin{center}
\includegraphics[width=1\linewidth]{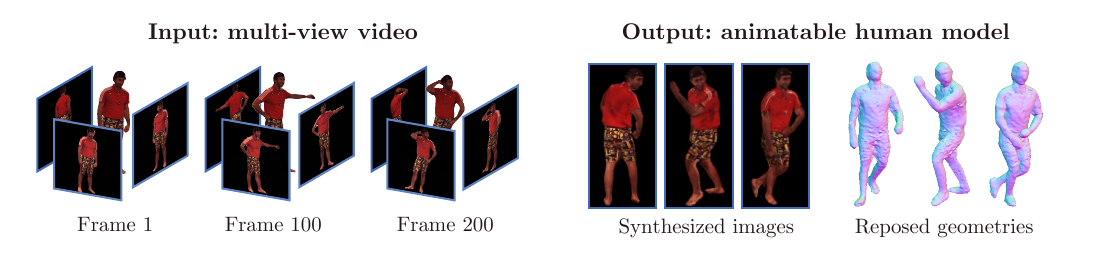}
\end{center} \vspace{-1.5em}
\captionof{figure}{Given a multi-view video of a performer, our method reconstructs an animatable human model, which can be used for novel view synthesis and 3D shape generation under novel human poses.}
\label{fig:teaser}
\bigbreak]

\ificcvfinal\thispagestyle{empty}\fi

\begin{abstract}
    This paper addresses the challenge of reconstructing an animatable human model from a multi-view video. Some recent works have proposed to decompose a non-rigidly deforming scene into a canonical neural radiance field and a set of deformation fields that map observation-space points to the canonical space, thereby enabling them to learn the dynamic scene from images. However, they represent the deformation field as translational vector field or SE(3) field, which makes the optimization highly under-constrained. Moreover, these representations cannot be explicitly controlled by input motions. Instead, we introduce neural blend weight fields to produce the deformation fields. Based on the skeleton-driven deformation, blend weight fields are used with 3D human skeletons to generate observation-to-canonical and canonical-to-observation correspondences. Since 3D human skeletons are more observable, they can regularize the learning of deformation fields. Moreover, the learned blend weight fields can be combined with input skeletal motions to generate new deformation fields to animate the human model. Experiments show that our approach significantly outperforms recent human synthesis methods. The code and supplementary materials are available at \href{https://zju3dv.github.io/animatable\_nerf/}{https://zju3dv.github.io/animatable\_nerf/}.
\end{abstract}

\let\thefootnote\relax\footnotetext{$^*$The first two authors contributed equally. The authors from Zhejiang University are affiliated with the State Key Lab of CAD\&CG.}
\let\thefootnote\relax\footnotetext{$^{\dagger}$Corresponding author: Hujun Bao.}
\section{Introduction}


Rendering animatable human characters has a variety of applications such as free-viewpoint videos, telepresence, video games and movies. 
The core step is to reconstruct animatable human models, which tends to be expensive and time-consuming in traditional pipelines due to two factors.
First, high-quality human reconstruction generally relies on complicated hardware, such as a dense array of cameras \cite{schonberger2016structure, guo2019relightables} or depth sensors \cite{collet2015high, dou2016fusion4d}. Second, human animation requires skilled artists to manually create a skeleton suitable for the human model and carefully design skinning weights \cite{lewis2000pose} to achieve realistic animation, which takes countless human labor.


In this work, we aim to reduce the cost of human reconstruction and animation, to enable the creation of digital humans at scale. Specifically, we focus on the problem of automatically reconstructing animatable humans from multi-view videos, as illustrated in Figure~\ref{fig:teaser}. 
However, this problem is extremely challenging. There are two core questions we need to answer: how to represent animatable human models and how to learn this representation from videos?


Recently, neural radiance fields (NeRF) \cite{mildenhall2020nerf} has proposed a representation that can be efficiently learned from images with a differentiable renderer. It represents static 3D scenes as color and density fields, which work particularly well with volume rendering techniques. To extend NeRF to handle non-rigidly deforming scenes, \cite{park2020deformable, pumarola2020d} decompose a video into a canonical NeRF and a set of deformation fields that transform observation-space points at each video frame to the canonical space. The deformation field is represented as translational vector field \cite{pumarola2020d} or SE(3) field \cite{park2020deformable}. 
Although they can handle some dynamic scenes, they are not suited for representing animatable human models due to two reasons. First, jointly optimizing NeRF with translational vector fields or SE(3) fields without motion prior is an extremely under-constrained problem \cite{pumarola2020d, li2020neural}. Second, they cannot explicitly synthesize novel scenes given input motions for animation.

To overcome these problems, we propose a novel motion representation named neural blend weight field. Based on the skeleton-driven deformation framework \cite{lewis2000pose}, blend weight fields are combined with 3D human skeletons to generate deformation fields. This representation has two advantages. First, since the human skeleton is easy to track \cite{joo2018total}, it does not need to be jointly optimized and thus provides an effective regularization on the learning of deformation fields. Second, by learning an additional neural blend weight field at the canonical space, we can explicitly animate the neural radiance field with input motions.


We evaluate our approach on the H36M \cite{ionescu2013human3} and ZJU-MoCap \cite{peng2020neural} datasets that capture dynamic humans in complex motions with synchronized cameras. Across all video sequences, our approach exhibits state-of-the-art performances on novel view synthesis and novel pose synthesis.
In addition, our method is able to reconstruct the 3D human shape at the canonical space and repose the geometry.


In summary, this work has the following contributions:

\begin{itemize}
    \item We introduce a novel representation called neural blend weight field, which can be combined with NeRF and 3D human skeletons to recover animatable human models from multi-view videos.
    \item Our approach demonstrates significant performance improvement on novel view synthesis and novel pose synthesis compared to recent human synthesis methods on the H36M and ZJU-MoCap datasets.
\end{itemize}

\vspace{-0.5em}
\begin{figure*}[t]
\centering
\includegraphics[width=1\linewidth]{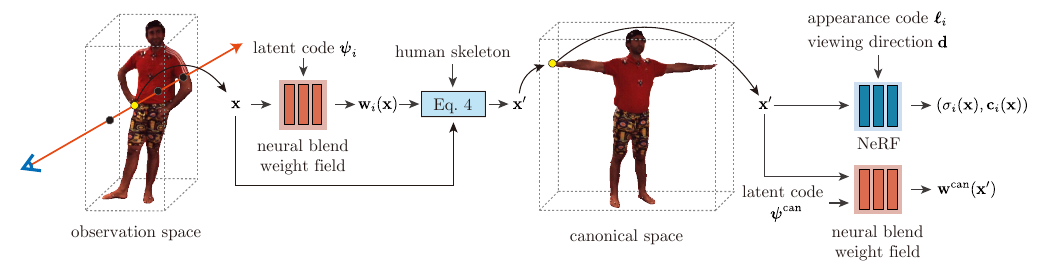}
\vspace{-2em}
\caption{\textbf{Overview of our approach.} Given a query point $\mathbf{x}$ in the observation space at frame $i$, we infer its blend weight $\mathbf{w}_i(\mathbf{x})$ using a neural blend weight field that is conditioned on the latent code $\boldsymbol{\psi}_i$. Based on the blend weight and the human skeleton, we can obtain the corresponding point $\mathbf{x}'$ in the canonical space using equation (\ref{eq:inverse}). Taking the transformed point $\mathbf{x}'$, observation-space viewing direction $\mathbf{d}$, and appearance code $\boldsymbol{\ell}_i$ as inputs, the template NeRF model predicts the volume density and color. To animate the template NeRF, we also learn a neural blend field $\mathbf{w}^{\text{can}}(\mathbf{x}')$ at the canonical space.}
\label{fig:pipeline}
\end{figure*}

\section{Related work}


\noindent \textbf{Human reconstruction.} Modeling human characters is the first step of traditional animation pipelines. To achieve high-quality reconstruction, most methods rely on complicated hardware \cite{collet2015high, dou2016fusion4d, su2020robustfusion, debevec2000acquiring, guo2019relightables}. Recently, some works \cite{sitzmann2019scene, niemeyer2020differentiable, mildenhall2020nerf, yariv2020multiview} have attempted to learn 3D representations from images with differentiable renderers, which reduces the number of input camera views and achieves impressive reconstruction results. However, they have difficulty in recovering reasonable 3D human shapes when the camera views are too sparse, as shown in \cite{peng2020neural}. Instead of optimizing the network parameters per scene, \cite{natsume2019siclope, saito2019pifu, zheng2019deephuman, saito2020pifuhd} utilize networks to learn human shape priors from ground-truth 3D data, allowing them to reconstruct human shapes from even a single image.


\noindent \textbf{Human animation.} Skeletal animation \cite{lewis2000pose, kavan2007skinning} is a common approach to animate human models. It first creates a scale-appropriate skeleton for the human mesh and then assigns each mesh vertex a blend weight that describes how the vertex position deforms with the skeleton. Skinned multi-person linear model (SMPL) \cite{loper2015smpl} learns a skeleton regressor and blend weights from a large amount of ground-truth 3D meshes. Based on SMPL, some works \cite{pavlakos2018learning, kanazawa2018end, kolotouros2019cmr, jiang2020mpshape, dong2020motion} reconstruct an animated human mesh from sparse camera views. However, SMPL only describes the naked human body and thus cannot be directly used to render photorealistic images. To overcome this problem, \cite{alldieck2018video, alldieck2019learning, alldieck2019tex2shape} apply vertex displacements to the SMPL model to capture the human clothing and hair. \cite{weng2019photo} proposes a 2D warping method to deform the SMPL model to fit the input image. Recent implicit function-based methods \cite{park2019deepsdf, mescheder2019occupancy, chibane2020implicit} have exhibited state-of-the-art reconstruction quality. \cite{huang2020arch, bhatnagar2020ipnet} combine implicit function learning with the SMPL model to obtain detailed animatable human models. \cite{deng2020nasa} combines a set of local implicit functions with human skeletons to represent dynamic humans. \cite{yang2021s3} proposes to animate occupancy networks with a linear blend skinning algorithm. However, these methods all need the supervision of 3D ground-truth data.


\noindent \textbf{Neural rendering.} To reduce the requirement for the reconstruction quality, some methods \cite{shysheya2019textured, thies2019deferred, liu2020neural, wu2020multi, kwon2020rotationally} improve the rendering pipeline with neural networks. Based on the advances in image-to-image translation techniques \cite{isola2017image}, \cite{ma2017pose, chan2019everybody, men2020controllable} train a network to map 2D skeleton images to target rendering results. Although these methods can synthesize photorealistic images under novel human poses, they have difficulty in rendering novel views. To improve the performance of novel view synthesis, \cite{shysheya2019textured, thies2019deferred, wu2020multi, aliev2020neural, prokudin2021smplpix, yoon2021pose, raj2020anr} introduce 3D representations into the rendering pipeline. \cite{thies2019deferred} establishes neural texture maps and uses UV maps to obtain feature maps in the image space, which is then interpreted into images with a neural renderer. \cite{wu2020multi, aliev2020neural} reconstruct a point cloud from input images and learn a 3D feature for each point. Then, they project 3D features into a 2D feature map and employ a network to render images. However, 2D convolutional networks have difficulty in rendering inter-view consistent images, as shown in \cite{sitzmann2019scene}.

To solve this problem, \cite{lombardi2019neural, niemeyer2020differentiable, mildenhall2020nerf, li2020crowd, liu2020nsvf} interpret features into colors in 3D space and then accumulate them into 2D images. \cite{lombardi2019neural} uses 3D convolutional networks to produce discretized RGB-$\alpha$ volumes. Neural radiance fields (NeRF) \cite{mildenhall2020nerf} proposes to represent 3D scenes with color and density fields, which works well with the volumetric rendering and gives state-of-the-art performances on novel view synthesis. \cite{peng2020neural} combines NeRF with the SMPL model, allowing it to handle dynamic humans and synthesize photorealistic novel views from very sparse camera views.

\section{Method}


Given a multi-view video of a performer, our task is to reconstruct an animatable human model that can be used to synthesize free-viewpoint videos of the performer under novel human poses. The cameras are synchronized and calibrated. For each frame, we assume the 3D human skeleton is given, which can be obtained with marker-based or marker-less pose estimation systems \cite{ionescu2013human3, joo2018total}. For each image, \cite{gong2018instance} is used to extract the foreground human mask, and the values of the background image pixels are set as zero.

The overview of our approach is shown in Figure~\ref{fig:pipeline}. We decompose a non-rigidly deforming human body into a canonical human model represented by a neural radiance field (Section~\ref{section:nerf}) and a per-frame blend weight field (Section~\ref{section:nsf}) that is used to establish correspondences between the observation space and canonical space. Then we discuss how to learn the representation on the multi-view video (Section~\ref{section:optim}). Based on blend weight fields, we are able to animate the canonical human model (Section~\ref{section:anerf}).

\subsection{Representing videos with neural radiance fields}

\label{section:nerf}

NeRF represents a static scene as a continuous volumetric representation. For any 3D point, it takes a spatial position $\mathbf{x}$ and viewing direction $\mathbf{d}$ as input to a neural network and outputs a volume density $\sigma$ and color $\mathbf{c}$. 

Inspired by \cite{park2020deformable, pumarola2020d}, we extend NeRF to represent the dynamic human body by introducing deformation fields, as shown in Figure~\ref{fig:pipeline}. Specifically, for each video frame $i \in \{1, ..., N\}$, we define a deformation field $T_i$ that transforms observation-space points to the canonical space. Given the canonical-frame density model $F_{\sigma}$, the density model at frame $i$ can be thus defined as:
\begin{equation}
    (\sigma_i(\mathbf{x}), \mathbf{z}_i(\mathbf{x})) = F_{\sigma}(\gamma_\mathbf{x}(T_i(\mathbf{x}))),
    \label{eq:sigma}
\end{equation}
where $\mathbf{z}_i(\mathbf{x})$ is the shape feature in the original NeRF, and $\gamma_\mathbf{x}$ is the positional encoding \cite{mildenhall2020nerf} for spatial location.

When predicting the color, we define a per-frame latent code $\boldsymbol{\ell}_i$ to encode the state of the human appearance in frame $i$. Similarly, with the canonical-frame color model $F_{\mathbf{c}}$, the color model at frame $i$ can be defined as:
\begin{equation}
    \mathbf{c}_i(\mathbf{x}) = F_{\mathbf{c}}(\mathbf{z}_i(\mathbf{x}), \gamma_\mathbf{d}(\mathbf{d}), \boldsymbol{\ell}_i),
    \label{eq:color}
\end{equation}
where $\gamma_\mathbf{d}$ is the positional encoding for viewing direction.

There are several ways to represent the deformation field, such as translational vector field \cite{pumarola2020d, li2020neural} and SE(3) field \cite{park2020deformable}. However, as discussed in \cite{park2020deformable, li2020neural}, optimizing a radiance field together with a deformation field is an ill-posed problem that is prone to local optima. To overcome this problem, \cite{park2020deformable, li2020neural} propose many regularization techniques to facilitate the training, which makes the optimization process complex. Moreover, their representations cannot robustly generate new deformation fields given novel motion sequences.

\subsection{Neural blend weight fields}

\label{section:nsf}

Considering that we aim to model dynamic humans, it is natural to leverage the human priors to learn the deformation field, which helps us to solve the under-constrained problem. Specifically, we construct the deformation field based on the 3D human skeleton and the skeleton-driven deformation framework \cite{lewis2000pose}. 

The human skeleton defines $K$ parts, which produce $K$ transformation matrices $\{G_k\} \in SE(3)$. The detailed derivation is listed in the supplementary material. In the linear blend skinning algorithm \cite{lewis2000pose}, a canonical-space point $\mathbf{v}$ is transformed to the observation space using
\begin{equation}
    \mathbf{v}' = \left( \sum_{k=1}^K w(\mathbf{v})_k G_k \right) \mathbf{v},
    \label{eq:skinning}
\end{equation}
where $w(\mathbf{v})_k$ is the blend weight of $k$-th part. Similarly, for an observation-space point $\mathbf{x}$, if we know its corresponding blend weights, we are able to transform it to the canonical space using
\begin{equation}
    \mathbf{x}' = \left( \sum_{k=1}^K w^o(\mathbf{x})_k G_k \right)^{-1} \mathbf{x},
    \label{eq:inverse}
\end{equation}
where $w^o(\mathbf{x})$ is the blend weight function defined in the observation space. To obtain the blend weight field, a natural idea is to define a function that maps a 3D point to blend weights, which then gives the dynamic radiance fields based on equations \eqref{eq:sigma}, \eqref{eq:color} and \eqref{eq:inverse}. However, we find that jointly learning NeRF with the blend weight field is still ill-posed and is prone to local minima.

To solve this problem, we seek the human priors in 3D statistical body models \cite{loper2015smpl, romero2017embodied, pavlakos2019expressive, xu2020ghum} to regularize the learned blend weights. Specifically, for any 3D point, we assign an initial blend weight based on the body model and then use a network to learn a residual vector, resulting in the neural blend weight field. In practice, the residual vector fields for all training video frames are implemented using a single MLP network $F_{\Delta \mathbf{w}}: (\mathbf{x}, \boldsymbol{\psi}_i) \rightarrow \Delta \mathbf{w}_i$, where $\boldsymbol{\psi}_i$ is a per-frame learned latent code and $\Delta \mathbf{w}_i$ is a vector $\in \mathbb{R}^{K}$. The neural blend weight field at frame $i$ is defined as:
\begin{equation}
    \mathbf{w}_i(\mathbf{x}) = \text{norm}(F_{\Delta \mathbf{w}}(\mathbf{x}, \boldsymbol{\psi}_i) + \mathbf{w}^{\text{s}}(\mathbf{x}, S_i)),
\end{equation}
where $\mathbf{w}^{\text{s}}$ is the initial blend weights that are computed based on the statistical body model $S_i$, and we define $\text{norm}(\mathbf{w})=\mathbf{w}/ \sum w_i$. Without loss of generality, we adopt SMPL \cite{loper2015smpl} as the body model, which can be obtained by fitting the SMPL model to the 3D human skeleton \cite{joo2018total}. Note that this idea can also apply to other human models \cite{romero2017embodied, pavlakos2019expressive, xu2020ghum}. To compute $\mathbf{w}^{\text{s}}$, we take the strategy proposed in \cite{huang2020arch, bhatnagar2020loopreg}. For any 3D point, we first find the closest surface point on the SMPL mesh. Then, the target blend weight is computed by performing barycentric interpolation of the blend weights of three vertices on the corresponding mesh facet.

To animate the learned template NeRF, we additionally learn a neural blend weight field $\mathbf{w}^{\text{can}}$ at the canonical space. The SMPL blend weight field $\mathbf{w}^{\text{s}}$ is calculated using the T-pose SMPL model, and $F_{\Delta \mathbf{w}}$ is conditioned on an additional latent code $\boldsymbol{\psi^{\text{can}}}$. We utilize the inherent consistency between blend weights to optimize the neural blend weight field $\mathbf{w}^{\text{can}}$, which will be described in Section~\ref{section:optim}.

Instead of learning blend weight fields at both observation and canonical spaces, an alternative method is to only learn the blend weight field at the canonical space as in Equation \eqref{eq:skinning}, which specifies the canonical-to-observation correspondences. However, ``inverting'' Equation \eqref{eq:skinning} to get observation-to-canonical correspondences for rendering is non-trivial. We would need to first build a dense set of observation-to-canonical correspondences by densely sampling points at the canonical space and evaluating their blend weights. Then, for any observation-space point, we can interpolate its corresponding canonical point based on the pre-computed correspondences. This process is complex and time-consuming. Moreover, as the sampled points are discretized, the calculated correspondences tend to be coarse. In contrast, learning blend weights at observation spaces enables us to easily obtain the observation-to-canonical correspondences based on Equation \eqref{eq:inverse}.

\subsection{Training}

\label{section:optim}

Based on the dynamic radiance field $\sigma_i$ and $\mathbf{c}_i$, we can use volume rendering techniques \cite{kajiya1986rendering, mildenhall2020nerf} to synthesize images of particular viewpoints for each video frame $i$. The near and far bounds of volume rendering are estimated by computing the 3D boxes that bound the SMPL meshes. The parameters of $F_{\sigma}$, $F_{\mathbf{c}}$, $F_{\Delta \mathbf{w}}$, $\{\boldsymbol{\ell}_i\}$ and $\{\boldsymbol{\psi}_i\}$ are jointly optimized over the multi-view video by minimizing the difference between the rendered pixel color $\tilde{\mathbf{C}}_i(r)$ and the observed pixel color $\mathbf{C}_i(r)$:
\begin{equation}
    L_{\text{rgb}} = \sum_{r \in \mathcal{R}} \|\tilde{\mathbf{C}}_i(\mathbf{r}) - \mathbf{C}_i(\mathbf{r}) \|_2,
\end{equation}
where $\mathcal{R}$ is the set of rays passing through image pixels.

To learn the neural blend weight field $\mathbf{w}^{\text{can}}$ at the canonical space, we introduce a consistency loss between blend weight fields. As shown by equations \eqref{eq:skinning} and \eqref{eq:inverse}, two corresponding points at canonical and observation spaces should have the same blend weights. For an observation-space point $\mathbf{x}$ at frame $i$, we map it to the canonical-space point $T_i(\mathbf{x})$ using equation \eqref{eq:inverse}. The consistency loss between blend weight fields is defined as:
\begin{equation}
    L_{\text{nsf}} = \sum_{\mathbf{x} \in \mathcal{X}_i} \| \mathbf{w}_i(\mathbf{x}) - \mathbf{w}^{\text{can}}(T_i(\mathbf{x})) \|_1,
\end{equation}
where $\mathcal{X}_i$ is the set of 3D points sampled within the 3D human bounding box at frame $i$. The coefficient weights of $L_{\text{rgb}}$ and $L_{\text{nsf}}$ are both set to 1.

\subsection{Animation}

\label{section:anerf}

\paragraph{Image synthesis.} To synthesize images of the performer under novel human poses, we similarly construct the deformation fields that transform the 3D points to the canonical space. Given a novel human pose, our method updates the pose parameters in the SMPL model and computes the SMPL blend weight field $\mathbf{w}^{\text{s}}$ based on the new parameters $S^{\text{new}}$. Then, the neural blend weight field $\mathbf{w}^{\text{new}}$ for the novel human pose is defined as:
\begin{equation}
    \resizebox{.9\hsize}{!}{$\mathbf{w}^{\text{new}}(\mathbf{x}, \boldsymbol{\psi}^{\text{new}}) = \text{norm}(F_{\Delta \mathbf{w}}(\mathbf{x}, \boldsymbol{\psi}^{\text{new}})  +  \mathbf{w}^{\text{s}}(\mathbf{x}, S^{\text{new}})),$}
\end{equation}
where the $F_{\Delta \mathbf{w}}$ is conditioned on a new latent code $\boldsymbol{\psi}^{\text{new}}$. Based on the $\mathbf{w}^{\text{new}}$ and equation \eqref{eq:inverse}, we can generate the deformation field $T^{\text{new}}$ for the novel human pose. The parameters of $\boldsymbol{\psi}^{\text{new}}$ are optimized using
\begin{equation}
    L_{\text{new}} = \sum_{\mathbf{x} \in \mathcal{X}^{\text{new}}} \| \mathbf{w}^{\text{new}}(\mathbf{x}) - \mathbf{w}^{\text{can}}(T^{\text{new}}(\mathbf{x})) \|_1,
    \label{eq:lnew}
\end{equation}
where $\mathcal{X}^{\text{new}}$ is the set of 3D points sampled within the human box under the novel human pose. Note that we fix the parameters of $\mathbf{w}^{\text{can}}$ during training.  In practice, we train neural skinning fields under multiple novel human poses simultaneously. This is implemented by conditioning $F_{\Delta \mathbf{w}}$ on multiple latent codes. With the deformation field $T^{\text{new}}$, our method uses equations \eqref{eq:sigma} and \eqref{eq:color} to produce the neural radiance field under the novel human pose. 


\paragraph{3D shape generation.} In addition to synthesizing images under novel human poses, our approach can also explicitly animate a reconstructed human mesh, similar to the traditional animation methods. In particular, we first discretize the human bounding box at the canonical space with a voxel size of $5mm \times 5mm \times 5mm$ and evaluate the volume densities for all voxels, which are used to extract the human mesh with the Marching Cubes algorithm \cite{lorensen1987marching}. Then, blend weights of mesh vertices are inferred from the neural blend weight field $\mathbf{w}^{\text{can}}$. Finally, given a novel human pose, we use equation \eqref{eq:skinning} to transform each vertex, resulting in a deformed mesh under the target pose. The reconstruction results are presented in the supplementary material.

\section{Implementation details}

The networks of our radiance field $F_{\sigma}$ and $F_{\mathbf{c}}$ closely follow the original NeRF \cite{mildenhall2020nerf}. We only use the single-level NeRF and sample 64 points along each camera ray. The network of $F_{\Delta \mathbf{w}}$ is almost the same as that of $F_{\sigma}$, except that the final output layer of $F_{\Delta \mathbf{w}}$ has 24 channels. In addition, $F_{\Delta \mathbf{w}}$ applies $\exp(\cdot)$ to the output. The details of network architectures are described in the supplementary material. The appearance code $\boldsymbol{\ell}_i$ and blend weight field code $\boldsymbol{\psi}_i$ both have dimensions of 128.

\textbf{Training.} Our method takes a two-stage training pipeline. First, we train the parameters of $F_{\sigma}$, $F_{\mathbf{c}}$, $F_{\Delta \mathbf{w}}$, $\{\boldsymbol{\ell}_i\}$ and $\{\boldsymbol{\psi}_i\}$ jointly over the input video. Second, neural blend weight fields under novel human poses are learned using equation \eqref{eq:lnew}. The Adam optimizer \cite{kingma2014adam} is adopted for the training. The learning rate starts from $5e^{-4}$ and decays exponentially to $5e^{-5}$ along the optimization. The training is conducted on four 2080 Ti GPUs. For a three-view video of 300 frames, the first stage training takes around 200k iterations to converge (about 12 hours). For 200 novel human poses, the second stage training takes around 10k iterations to converge (about 30 minutes).

\section{Experiments}

\subsection{Dataset and metrics}

\noindent \textbf{H36M \cite{ionescu2013human3}} records multi-view videos with 4 cameras and collects human poses using the marker-based motion capture system. It includes multiple subjects performing complex actions. We select representative actions, split the videos into training and test frames, and perform experiments on subjects S1, S5, S6, S7, S8, S9, and S11. Three cameras are used for training and the remaining camera is selected for test. We use \cite{joo2018total} to obtain the SMPL parameters from the 3D human poses and apply \cite{gong2018instance} to segment foreground humans. More details of training and test data can be found in the supplementary material.

\noindent \textbf{ZJU-MoCap \cite{peng2020neural}} records multi-view videos with 21 cameras and collects human poses using the marker-less motion capture system. For evaluation, we select four representative sequences: ``Twirl", ``Taichi", ``Warmup", and ``Punch1". Four uniformly distributed cameras are used for training and the remaining cameras for testing. We follow the experimental protocol in \cite{peng2020neural}.



\noindent \textbf{Metrics.} Following typical protocols \cite{mildenhall2020nerf}, we evaluate our method on image synthesis using two metrics: peak signal-to-noise ratio (PSNR) and structural similarity index (SSIM). For 3D reconstruction, since there is no ground-truth geometry, we only provide qualitative results, which can be found in the supplementary material.

\subsection{Performance on image synthesis}

\paragraph{Baselines.} We compare with state-of-the-art image synthesis methods \cite{thies2019deferred, wu2020multi, peng2020neural} that also utilize SMPL priors. 1) Neural Textures \cite{thies2019deferred} renders a coarse mesh with latent texture maps and uses a 2D CNN to interpret feature maps into target images. Since \cite{thies2019deferred} is not open-sourced, we reimplement it and take the SMPL mesh as the input mesh. 2) NHR \cite{wu2020multi} extracts 3D features from input point clouds and renders them into 2D feature maps, which are then transformed into images using 2D CNNs. Since dense point clouds are difficult to obtain from sparse camera views, we take SMPL vertices as input point clouds. 3) Neural body \cite{peng2020neural} represents the human body with an implicit field conditioned on the latent codes anchored on the vertices of SMPL and renders the images using volume rendering.

\begin{figure*}[t]
\centering
\includegraphics[width=1\linewidth]{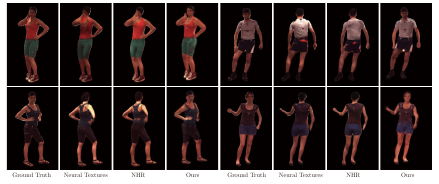}
\vspace{-2em}
\caption{\textbf{Qualitative results of novel view synthesis on the H36M dataset.} \cite{thies2019deferred, wu2020multi} have difficulty in controlling the viewpoint and seem to overfit training views. Compared with them, our method accurately renders the target view.}
\vspace{-1em}
\label{fig:h36m_view}
\end{figure*}

\paragraph{Results of novel view synthesis.} For comparison, we synthesize novel views of training video frames. Table~\ref{table:h36m_view_result} shows the comparison of our method with \cite{thies2019deferred, wu2020multi}. Specifically, our model outperforms \cite{thies2019deferred, wu2020multi} by a margin of at least 2.07 in terms of the PSNR metric and 0.024 in terms of the SSIM metric. Moreover, the proposed method achieves comparable results with the most recent state-of-the-art approach \cite{peng2020neural} as shown in Table~\ref{table:zjum_view_result}, despite not being specifically designed for the novel view synthesis task.

\begin{table}
\begin{center}
\scalebox{0.92}{
\tablestyle{4pt}{1.05}
\begin{tabular}{c|x{26}x{33}x{25}|x{26}x{33}x{25}}
& \multicolumn{3}{c|}{PSNR} & \multicolumn{3}{c}{SSIM} \\[.1em]
\shline
& NT \cite{thies2019deferred} & NHR \cite{wu2020multi} & Ours & NT \cite{thies2019deferred} & NHR \cite{wu2020multi} & Ours \\
\hline
S1  & 20.98 & 21.08 & \textbf{22.05} & 0.860 & 0.872 & \textbf{0.888} \\
S5  & 19.87 & 20.64 & \textbf{23.27} & 0.855 & 0.872 & \textbf{0.892} \\
S6  & 20.18 & 20.40 & \textbf{21.13} & 0.816 & 0.830 & \textbf{0.854} \\
S7  & 20.47 & 20.29 & \textbf{22.50} & 0.856 & 0.868 & \textbf{0.890} \\
S8  & 16.77 & 19.13 & \textbf{22.75} & 0.837 & 0.871 & \textbf{0.898} \\
S9  & 22.96 & 23.04 & \textbf{24.72} & 0.873 & 0.879 & \textbf{0.908} \\
S11 & 21.71 & 21.91 & \textbf{24.55} & 0.859 & 0.871 & \textbf{0.902} \\
\hline                                        
average & 20.42 & 20.93 & \textbf{23.00} & 0.851 & 0.866 & \textbf{0.890} \\
\end{tabular}
}
\end{center}
\vspace{-1.5em}
\caption{\textbf{Results of novel view synthesis on H36M dataset in terms of PSNR and SSIM (higher is better).} ``NT" means Neural Textures.}
\vspace{-1em}
\label{table:h36m_view_result}
\end{table}


\begin{table}
\begin{center}
\scalebox{0.86}{
\tablestyle{4pt}{1.05}
\begin{tabular}{x{35}|x{20}x{20}x{20}x{20}|x{20}x{20}x{20}x{20}}
& \multicolumn{4}{c|}{PSNR} & \multicolumn{4}{c}{SSIM} \\[.1em]
\shline
& NT \cite{thies2019deferred} & NHR \cite{wu2020multi} & NB \cite{peng2020neural} & Ours & NT \cite{thies2019deferred} & NHR \cite{wu2020multi} & NB \cite{peng2020neural} & Ours \\
\hline     
  
novel view	& 22.61 	& 23.25 	& \textbf{28.90} 	& 27.10 	& 0.899 	& 0.905 	& \textbf{0.967} 	& 0.949 \\
novel pose	& 21.55 	& 21.88 	& 23.06 	& \textbf{23.16} 	& 0.860 	& 0.863 	& 0.879 	& \textbf{0.893} \\
\end{tabular}
}
\end{center}
\vspace{-1.5em}
\caption{\textbf{Results of novel view synthesis and novel pose synthesis on ZJU-MoCap dataset in terms of PSNR and SSIM (higher is better).} ``NB'' means Neural Body.}
\vspace{-1.5em}
\label{table:zjum_view_result}
\end{table}

Figure \ref{fig:h36m_view} presents the qualitative comparison of our method with \cite{thies2019deferred, wu2020multi}. Both \cite{thies2019deferred, wu2020multi} have difficulty in controlling the rendering viewpoint and tend to synthesize contents of training views. As shown in the second person of Figure~\ref{fig:h36m_view}, they render the human back that is seen during training. In contrast, our method is able to accurately control the viewpoint, thanks to the explicit 3D representation.



\begin{figure*}[t]
\centering
\includegraphics[width=1\linewidth]{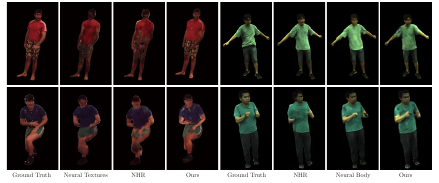}
\vspace{-2em}
\caption{\textbf{Qualitative results of novel pose synthesis on the H36M and ZJU-MoCap datasets.} For complex human poses, \cite{thies2019deferred, wu2020multi, peng2020neural} tend to generate distorted rendering results. In contrast to them, our method has a better generalization ability.}
\label{fig:h36m}
\end{figure*}

\begin{table}
\begin{center}
\scalebox{0.92}{
\tablestyle{4pt}{1.05}
\begin{tabular}{c|x{26}x{33}x{25}|x{26}x{33}x{25}}
& \multicolumn{3}{c|}{PSNR} & \multicolumn{3}{c}{SSIM} \\[.1em]
\shline
& NT \cite{thies2019deferred} & NHR \cite{wu2020multi} & Ours & NT \cite{thies2019deferred} & NHR \cite{wu2020multi} & Ours \\
\hline
S1  & 20.09 & 20.48 & \textbf{21.37} & 0.837 & 0.853 & \textbf{0.868} \\
S5  & 20.03 & 20.72 & \textbf{22.29} & 0.843 & 0.860 & \textbf{0.875} \\
S6  & 20.42 & 20.47 & \textbf{22.59} & 0.844 & 0.856 & \textbf{0.884} \\
S7  & 20.03 & 19.66 & \textbf{22.22} & 0.838 & 0.852 & \textbf{0.878} \\
S8  & 16.69 & 18.83 & \textbf{21.78} & 0.824 & 0.855 & \textbf{0.882} \\
S9  & 22.20 & 22.18 & \textbf{23.72} & 0.851 & 0.860 & \textbf{0.886} \\
S11 & 21.72 & 22.12 & \textbf{23.91} & 0.854 & 0.867 & \textbf{0.889} \\
\hline                                        
average & 20.17 & 20.64 & \textbf{22.55} & 0.842 & 0.858 & \textbf{0.880} \\
\end{tabular}
}
\end{center}
\vspace{-1.3em}
\caption{\textbf{Results of novel pose synthesis on H36M dataset in terms of PSNR and SSIM (higher is better).} ``NT" means Neural Textures.}
\vspace{-1.2em}
\label{table:h36m_result}
\end{table}

\vspace{-0.5em}

\vspace{-0.5em}

\paragraph{Results of novel pose synthesis.} For comparison, we synthesize test video frames from the test camera view. Table~\ref{table:h36m_result} compares our method with \cite{thies2019deferred, wu2020multi} in terms of the PSNR metric and the SSIM metric. For both metrics, our method gives the best performances. Table~\ref{table:zjum_view_result} shows that our model also outperforms \cite{peng2020neural} when generating images under novel human poses on ZJU-MoCap dataset.

The qualitative results are shown in Figure~\ref{fig:h36m}. For complex human poses, \cite{thies2019deferred, wu2020multi, peng2020neural} give blurry and distorted rendering results. In contrast, synthesized images of our method achieve better visual quality. The results indicate that our model has better controllability on the image generation process than CNN-based methods.

\subsection{Ablation studies}

We conduct ablation studies on one subject (S9) of the H36M \cite{ionescu2013human3} dataset in terms of the novel pose synthesis performance. First, to analyze the benefit of learning $F_{\Delta \mathbf{w}}$, we compare neural blend weight field with SMPL blend weight field. Then, to explore the influence of human pose accuracy, we estimate SMPL parameters from predicted human poses \cite{cao2018openpose, joo2018total} and perform training on these parameters. Finally, we explore the performances of our method under different numbers of video frames and camera views. Tables \ref{table:nsf_ssf}, \ref{table:human_pose}, \ref{table:video_length},  and \ref{table:view} summarize the results of ablation studies.


\begin{figure}[t]
\centering
\includegraphics[width=1\linewidth]{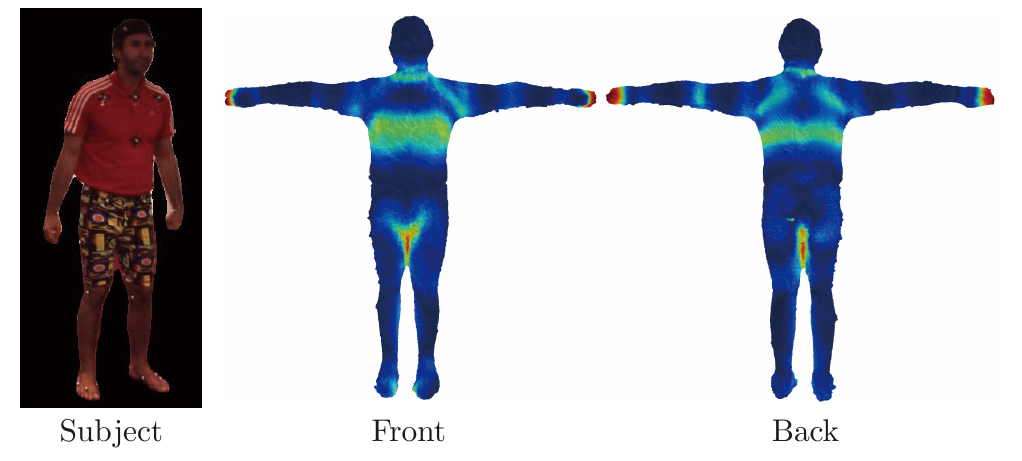}
\vspace{-2em}
\caption{\textbf{Visualization of the residual vector field $F_{\Delta \mathbf{w}}$} on the reconstructed geometries of subjects ``S9" and ``S6". Red means large residual. Best viewed in color.}
\label{fig:residual}
\end{figure}

\begin{table}
\begin{center}
\tablestyle{4pt}{1.05}
\begin{tabular}{x{90}|x{35}|x{35}}
& PSNR & SSIM \\[.1em]
\shline
Neural blend weight field & \textbf{23.72} & \textbf{0.886} \\
SMPL blend weight field & 21.65 & 0.850 \\
\end{tabular}
\end{center}
\vspace{-1em}
\caption{\textbf{Comparison between neural blend weight field and SMPL blend weight field} on subject ``S9".}
\vspace{-1em}
\label{table:nsf_ssf}
\end{table}

\textbf{Impact of neural blend weight field.} Table~\ref{table:nsf_ssf} shows the quantitative comparisons, which indicate that neural blend weight field performs better than SMPL blend weight field.

To better show the improvement on the SMPL blend weight field, Figure \ref{fig:residual} visualizes the residual vector field $F_{\Delta \mathbf{w}}$ on our reconstructed geometry at the canonical space. The bigger residual has a redder color. We can see that regions of big residual mainly locate on the neck, hand, chest, and pants, which are human-specific details that SMPL cannot describe. The results indicate that our learned $F_{\Delta \mathbf{w}}$ are physically interpretable.

\begin{table}
\begin{center}
\tablestyle{4pt}{1.05}
\begin{tabular}{x{100}|x{35}|x{35}}
& PSNR & SSIM \\[.1em]
\shline
Marker-based pose estimation & \textbf{23.72} & \textbf{0.886} \\
Marker-less pose estimation & 22.27 & 0.858 \\
\end{tabular}
\end{center}
\vspace{-1.5em}
\caption{\textbf{Comparison between models trained with human poses} from marker-based and marker-less pose estimation methods on subject ``S9".}
\label{table:human_pose}
\end{table}

\begin{table}
\begin{center}
\tablestyle{4pt}{1.05}
\begin{tabular}{x{35}|x{35}|x{35}|x{35}|x{35}}
Frames & 1 & 100 & 200 & 800 \\[.1em]
\shline
PSNR & 20.29 & 23.40 & \textbf{23.69} & 23.16 \\
SSIM & 0.849 & 0.881 & \textbf{0.883} & 0.875 \\
\end{tabular}
\end{center}
\vspace{-1.5em}
\caption{\textbf{Results of models trained with different numbers of video frames} on subject ``S9" of H36M dataset.}
\vspace{-1.3em}
\label{table:video_length}
\end{table}

\begin{figure}[t]
\centering
\includegraphics[width=1\linewidth]{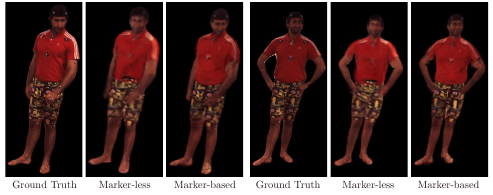}
\vspace{-1.8em}
\caption{\textbf{Qualitative results of models trained on poses} from marker-less and marker-based systems.}
\label{fig:human_pose}
\end{figure}

\textbf{Impact of the human pose accuracy.} Table \ref{table:human_pose} compares the models trained with human poses from marker-based and marker-less systems. The results show that more accurate human poses produce better rendering quality. The qualitative comparison is presented in Figure \ref{fig:human_pose}.

\begin{figure}[t]
\centering
\includegraphics[width=1\linewidth]{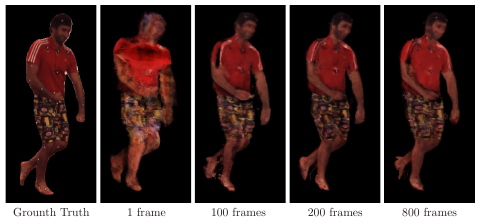}
\vspace{-1.8em}
\caption{\textbf{Comparison of models trained with different numbers of video frames} on the subject ``S9".}
\label{fig:video_length}
\end{figure}

\textbf{Impact of the video length.} For comparison, we take 1, 100, 200 and 800 video frames for training and test the models on the same motion sequence. Table~\ref{table:video_length} lists the quantitative results of our models trained with different numbers of video frames. The results demonstrate that training on the video helps the representation learning, but the network seems to have difficulty in fitting very long videos. Empirically, we find that 150$\sim$300 frames are suitable for most subjects. Figure~\ref{fig:video_length} presents the qualitative comparisons.

\begin{figure}[t]
\centering
\includegraphics[width=1\linewidth]{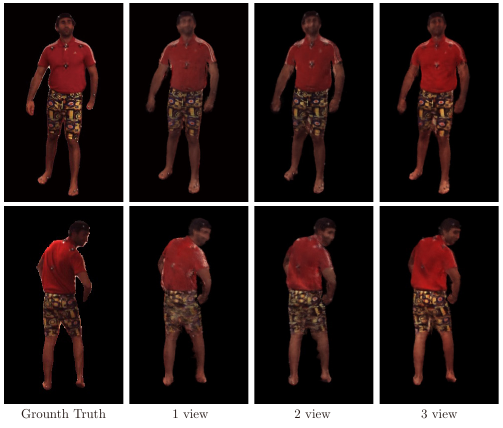}
\vspace{-1.8em}
\caption{\textbf{Comparison of models trained with different numbers of camera views} on the subject ``S9".}
\label{fig:view}
\end{figure}

\textbf{Impact of the number of input views.} For comparison, we take one view for test and select 1, 2, and 3 nearest views for training. Table \ref{table:view} compares the performances of models trained with different numbers of input views. Surprisingly, the three models have similar quantitative performances. Figure \ref{fig:view} further compares the three models, which shows that the model trained on 3 views renders more details. It is worth noting that the model trained on a single view already achieves reasonable rendering quality.

\begin{table}
\begin{center}
\tablestyle{4pt}{1.05}
\begin{tabular}{x{35}|x{35}|x{35}|x{35}}
& 1 view & 2 views & 3 views \\[.1em]
\shline
PSNR & 23.81 & \textbf{24.16} & 23.72 \\
SSIM & 0.877 & 0.880 & \textbf{0.886} \\
\end{tabular}
\end{center}
\vspace{-1.3em}
\caption{\textbf{Results of models trained with different numbers of camera views} on subject ``S9".}
\label{table:view}
\end{table}

\subsection{Running time}

For $512 \times 512$ images, our algorithm takes 1.09s to render an image on a desktop with an Intel i7 3.7GHz CPU and a GTX 1080 Ti GPU. Specifically, our implementation takes 0.39s for predicting the color and density fields, 0.63s for predicting the blend weight fields, and 0.07s for volume rendering. Because the number of points sampled along the ray is only 64 and the scene bound of a human is small, the rendering speed of our method is relatively fast.

\section{Limitations}

\label{section:limitations}

Combining neural radiance fields with blend weight fields enables us to obtain impressive performances on novel view synthesis and novel pose synthesis. However, our method has a few limitations. 1) The skeleton-driven deformation model \cite{lewis2000pose} cannot express the complex non-rigid deformations of garments. As a result, the performance of our method tends to degrade when reconstructing performers that wear loose clothes. It would be interesting to augment neural radiance fields with the deformation graph \cite{newcombe2015dynamicfusion} that can model local garment deformations. 2) Currently our method requires rather accurate 3D human skeletons. We hope that, in the future, we can find a way to refine human poses during training. 3) Same to NeRF, our proposed model is trained per-scene, which requires a lot of time to produce animatable human models. Generalizing the networks across different videos and reducing training time is left as future work. 4) Moreoever, the rendering time of our model is a bit high. It is could be solved with recent caching-based techniques \cite{yu2021plenoctrees, hedman2021snerg}.

\section{Conclusion}

We introduced a novel dynamic human representation for modeling animatable human characters from multi-view videos. Our method augments a neural radiance field with deformation fields that transform observation-space points to the canonical space. The deformation fields are constructed based on the skeleton-driven deformation framework, where we learn neural blend weight fields to generate observation-to-canonical and canonical-to-observation correspondences. The animatable neural radiance field is learned over the multi-view video with volume rendering and the consistency among blend weight fields. After training, our method can synthesize free-viewpoint videos of a performer given novel motion sequences. Experiments on the H36M and ZJU-MoCap datasets demonstrated that the proposed model achieves state-of-the-art performances on image synthesis under novel views and novel human poses.

\vspace{1em}
\noindent\textbf{Acknowledgements:} 
The authors from Zhejiang University would like to acknowledge the support from the National Key Research and Development Program of China (No. 2020AAA0108901) and NSFC (No. 62172364).

{\small
\bibliographystyle{ieee_fullname}
\bibliography{egbib}

\begin{thebibliography}{10}\itemsep=-1pt

\bibitem{aliev2020neural}
Kara-Ali Aliev, Artem Sevastopolsky, Maria Kolos, Dmitry Ulyanov, and Victor
  Lempitsky.
\newblock Neural point-based graphics.
\newblock In {\em ECCV}, 2020.

\bibitem{alldieck2019learning}
Thiemo Alldieck, Marcus Magnor, Bharat~Lal Bhatnagar, Christian Theobalt, and
  Gerard Pons-Moll.
\newblock Learning to reconstruct people in clothing from a single {RGB}
  camera.
\newblock In {\em CVPR}, 2019.

\bibitem{alldieck2018video}
Thiemo Alldieck, Marcus Magnor, Weipeng Xu, Christian Theobalt, and Gerard
  Pons-Moll.
\newblock Video based reconstruction of 3d people models.
\newblock In {\em CVPR}, 2018.

\bibitem{alldieck2019tex2shape}
Thiemo Alldieck, Gerard Pons-Moll, Christian Theobalt, and Marcus Magnor.
\newblock Tex2shape: Detailed full human body geometry from a single image.
\newblock In {\em ICCV}, 2019.

\bibitem{bhatnagar2020ipnet}
Bharat~Lal Bhatnagar, Cristian Sminchisescu, Christian Theobalt, and Gerard
  Pons-Moll.
\newblock Combining implicit function learning and parametric models for 3d
  human reconstruction.
\newblock In {\em ECCV}, 2020.

\bibitem{bhatnagar2020loopreg}
Bharat~Lal Bhatnagar, Cristian Sminchisescu, Christian Theobalt, and Gerard
  Pons-Moll.
\newblock Loopreg: Self-supervised learning of implicit surface
  correspondences, pose and shape for 3d human mesh registration.
\newblock In {\em NeurIPS}, 2020.

\bibitem{cao2018openpose}
Zhe Cao, Gines Hidalgo, Tomas Simon, Shih-En Wei, and Yaser Sheikh.
\newblock Openpose: realtime multi-person 2d pose estimation using part
  affinity fields.
\newblock {\em PAMI}, 2018.

\bibitem{chan2019everybody}
Caroline Chan, Shiry Ginosar, Tinghui Zhou, and Alexei~A Efros.
\newblock Everybody dance now.
\newblock In {\em ICCV}, 2019.

\bibitem{chibane2020implicit}
Julian Chibane, Thiemo Alldieck, and Gerard Pons-Moll.
\newblock Implicit functions in feature space for 3d shape reconstruction and
  completion.
\newblock In {\em CVPR}, 2020.

\bibitem{collet2015high}
Alvaro Collet, Ming Chuang, Pat Sweeney, Don Gillett, Dennis Evseev, David
  Calabrese, Hugues Hoppe, Adam Kirk, and Steve Sullivan.
\newblock High-quality streamable free-viewpoint video.
\newblock {\em ACM TOG}, 2015.

\bibitem{debevec2000acquiring}
Paul Debevec, Tim Hawkins, Chris Tchou, Haarm-Pieter Duiker, Westley Sarokin,
  and Mark Sagar.
\newblock Acquiring the reflectance field of a human face.
\newblock In {\em SIGGRAPH}, 2000.

\bibitem{deng2020nasa}
Boyang Deng, JP Lewis, Timothy Jeruzalski, Gerard Pons-Moll, Geoffrey Hinton,
  Mohammad Norouzi, and Andrea Tagliasacchi.
\newblock Nasa: Neural articulated shape approximation.
\newblock In {\em ECCV}, 2020.

\bibitem{dong2020motion}
Junting Dong, Qing Shuai, Yuanqing Zhang, Xian Liu, Xiaowei Zhou, and Hujun
  Bao.
\newblock Motion capture from internet videos.
\newblock In {\em ECCV}, 2020.

\bibitem{dou2016fusion4d}
Mingsong Dou, Sameh Khamis, Yury Degtyarev, Philip Davidson, Sean~Ryan Fanello,
  Adarsh Kowdle, Sergio~Orts Escolano, Christoph Rhemann, David Kim, Jonathan
  Taylor, et~al.
\newblock Fusion4d: Real-time performance capture of challenging scenes.
\newblock {\em ACM TOG}, 2016.

\bibitem{gong2018instance}
Ke Gong, Xiaodan Liang, Yicheng Li, Yimin Chen, Ming Yang, and Liang Lin.
\newblock Instance-level human parsing via part grouping network.
\newblock In {\em ECCV}, 2018.

\bibitem{guo2019relightables}
Kaiwen Guo, Peter Lincoln, Philip Davidson, Jay Busch, Xueming Yu, Matt Whalen,
  Geoff Harvey, Sergio Orts-Escolano, Rohit Pandey, Jason Dourgarian, et~al.
\newblock The relightables: Volumetric performance capture of humans with
  realistic relighting.
\newblock {\em ACM TOG}, 2019.

\bibitem{hedman2021snerg}
Peter Hedman, Pratul~P. Srinivasan, Ben Mildenhall, Jonathan~T. Barron, and
  Paul Debevec.
\newblock Baking neural radiance fields for real-time view synthesis.
\newblock In {\em ICCV}, 2021.

\bibitem{huang2020arch}
Zeng Huang, Yuanlu Xu, Christoph Lassner, Hao Li, and Tony Tung.
\newblock Arch: Animatable reconstruction of clothed humans.
\newblock In {\em CVPR}, 2020.

\bibitem{ionescu2013human3}
Catalin Ionescu, Dragos Papava, Vlad Olaru, and Cristian Sminchisescu.
\newblock Human3.6m: Large scale datasets and predictive methods for 3d human
  sensing in natural environments.
\newblock {\em PAMI}, 2013.

\bibitem{isola2017image}
Phillip Isola, Jun-Yan Zhu, Tinghui Zhou, and Alexei~A Efros.
\newblock Image-to-image translation with conditional adversarial networks.
\newblock In {\em CVPR}, 2017.

\bibitem{jiang2020mpshape}
Wen Jiang, Nikos Kolotouros, Georgios Pavlakos, Xiaowei Zhou, and Kostas
  Daniilidis.
\newblock Coherent reconstruction of multiple humans from a single image.
\newblock In {\em CVPR}, 2020.

\bibitem{joo2018total}
Hanbyul Joo, Tomas Simon, and Yaser Sheikh.
\newblock Total capture: A 3d deformation model for tracking faces, hands, and
  bodies.
\newblock In {\em CVPR}, 2018.

\bibitem{kajiya1986rendering}
James~T Kajiya.
\newblock The rendering equation.
\newblock In {\em SIGGRAPH}, 1986.

\bibitem{kanazawa2018end}
Angjoo Kanazawa, Michael~J Black, David~W Jacobs, and Jitendra Malik.
\newblock End-to-end recovery of human shape and pose.
\newblock In {\em CVPR}, 2018.

\bibitem{kavan2007skinning}
Ladislav Kavan, Steven Collins, Ji{\v{r}}{\'\i} {\v{Z}}{\'a}ra, and Carol
  O'Sullivan.
\newblock Skinning with dual quaternions.
\newblock In {\em I3D}, 2007.

\bibitem{kingma2014adam}
Diederik~P Kingma and Jimmy Ba.
\newblock Adam: A method for stochastic optimization.
\newblock In {\em ICLR}, 2015.

\bibitem{kolotouros2019cmr}
Nikos Kolotouros, Georgios Pavlakos, and Kostas Daniilidis.
\newblock Convolutional mesh regression for single-image human shape
  reconstruction.
\newblock In {\em CVPR}, 2019.

\bibitem{kwon2020rotationally}
YoungJoong Kwon, Stefano Petrangeli, Dahun Kim, Haoliang Wang, Eunbyung Park,
  Viswanathan Swaminathan, and Henry Fuchs.
\newblock Rotationally-temporally consistent novel view synthesis of human
  performance video.
\newblock In {\em ECCV}, 2020.

\bibitem{lewis2000pose}
John~P Lewis, Matt Cordner, and Nickson Fong.
\newblock Pose space deformation: a unified approach to shape interpolation and
  skeleton-driven deformation.
\newblock In {\em SIGGRAPH}, 2000.

\bibitem{li2020neural}
Zhengqi Li, Simon Niklaus, Noah Snavely, and Oliver Wang.
\newblock Neural scene flow fields for space-time view synthesis of dynamic
  scenes.
\newblock In {\em CVPR}, 2021.

\bibitem{li2020crowd}
Zhengqi Li, Wenqi Xian, Abe Davis, and Noah Snavely.
\newblock Crowdsampling the plenoptic function.
\newblock In {\em ECCV}, 2020.

\bibitem{yariv2020multiview}
Yariv Lior, Kasten Yoni, Moran Dror, Galun Meirav, Atzmon Matan, Basri Ronen,
  and Lipman Yaron.
\newblock Multiview neural surface reconstruction by disentangling geometry and
  appearance.
\newblock In {\em NeurIPS}, 2020.

\bibitem{liu2020nsvf}
Lingjie Liu, Jiatao Gu, Kyaw~Zaw Lin, Tat-Seng Chua, and Christian Theobalt.
\newblock Neural sparse voxel fields.
\newblock In {\em NeurIPS}, 2020.

\bibitem{liu2020neural}
Lingjie Liu, Weipeng Xu, Marc Habermann, Michael Zollhoefer, Florian Bernard,
  Hyeongwoo Kim, Wenping Wang, and Christian Theobalt.
\newblock Neural human video rendering by learning dynamic textures and
  rendering-to-video translation.
\newblock {\em TVCG}, 2020.

\bibitem{lombardi2019neural}
Stephen Lombardi, Tomas Simon, Jason Saragih, Gabriel Schwartz, Andreas
  Lehrmann, and Yaser Sheikh.
\newblock Neural volumes: Learning dynamic renderable volumes from images.
\newblock In {\em SIGGRAPH}, 2019.

\bibitem{loper2015smpl}
Matthew Loper, Naureen Mahmood, Javier Romero, Gerard Pons-Moll, and Michael~J
  Black.
\newblock Smpl: A skinned multi-person linear model.
\newblock {\em ACM TOG}, 2015.

\bibitem{lorensen1987marching}
William~E Lorensen and Harvey~E Cline.
\newblock Marching cubes: A high resolution 3d surface construction algorithm.
\newblock In {\em SIGGRAPH}, 1987.

\bibitem{ma2017pose}
Liqian Ma, Xu Jia, Qianru Sun, Bernt Schiele, Tinne Tuytelaars, and Luc
  Van~Gool.
\newblock Pose guided person image generation.
\newblock In {\em NeurIPS}, 2017.

\bibitem{men2020controllable}
Yifang Men, Yiming Mao, Yuning Jiang, Wei-Ying Ma, and Zhouhui Lian.
\newblock Controllable person image synthesis with attribute-decomposed gan.
\newblock In {\em CVPR}, 2020.

\bibitem{mescheder2019occupancy}
Lars Mescheder, Michael Oechsle, Michael Niemeyer, Sebastian Nowozin, and
  Andreas Geiger.
\newblock Occupancy networks: Learning 3d reconstruction in function space.
\newblock In {\em CVPR}, 2019.

\bibitem{mildenhall2020nerf}
Ben Mildenhall, Pratul~P Srinivasan, Matthew Tancik, Jonathan~T Barron, Ravi
  Ramamoorthi, and Ren Ng.
\newblock Nerf: Representing scenes as neural radiance fields for view
  synthesis.
\newblock In {\em ECCV}, 2020.

\bibitem{natsume2019siclope}
Ryota Natsume, Shunsuke Saito, Zeng Huang, Weikai Chen, Chongyang Ma, Hao Li,
  and Shigeo Morishima.
\newblock Siclope: Silhouette-based clothed people.
\newblock In {\em CVPR}, 2019.

\bibitem{newcombe2015dynamicfusion}
Richard~A Newcombe, Dieter Fox, and Steven~M Seitz.
\newblock Dynamicfusion: Reconstruction and tracking of non-rigid scenes in
  real-time.
\newblock In {\em CVPR}, 2015.

\bibitem{niemeyer2020differentiable}
Michael Niemeyer, Lars Mescheder, Michael Oechsle, and Andreas Geiger.
\newblock Differentiable volumetric rendering: Learning implicit 3d
  representations without 3d supervision.
\newblock In {\em CVPR}, 2020.

\bibitem{park2019deepsdf}
Jeong~Joon Park, Peter Florence, Julian Straub, Richard Newcombe, and Steven
  Lovegrove.
\newblock Deepsdf: Learning continuous signed distance functions for shape
  representation.
\newblock In {\em CVPR}, 2019.

\bibitem{park2020deformable}
Keunhong Park, Utkarsh Sinha, Jonathan~T Barron, Sofien Bouaziz, Dan~B Goldman,
  Steven~M Seitz, and Ricardo-Martin Brualla.
\newblock Deformable neural radiance fields.
\newblock {\em arXiv preprint arXiv:2011.12948}, 2020.

\bibitem{pavlakos2019expressive}
Georgios Pavlakos, Vasileios Choutas, Nima Ghorbani, Timo Bolkart, Ahmed~AA
  Osman, Dimitrios Tzionas, and Michael~J Black.
\newblock Expressive body capture: 3d hands, face, and body from a single
  image.
\newblock In {\em CVPR}, 2019.

\bibitem{pavlakos2018learning}
Georgios Pavlakos, Luyang Zhu, Xiaowei Zhou, and Kostas Daniilidis.
\newblock Learning to estimate 3d human pose and shape from a single color
  image.
\newblock In {\em CVPR}, 2018.

\bibitem{peng2020neural}
Sida Peng, Yuanqing Zhang, Yinghao Xu, Qianqian Wang, Qing Shuai, Hujun Bao,
  and Xiaowei Zhou.
\newblock Neural body: Implicit neural representations with structured latent
  codes for novel view synthesis of dynamic humans.
\newblock In {\em CVPR}, 2021.

\bibitem{prokudin2021smplpix}
Sergey Prokudin, Michael~J Black, and Javier Romero.
\newblock Smplpix: Neural avatars from 3d human models.
\newblock In {\em WCCV}, 2021.

\bibitem{pumarola2020d}
Albert Pumarola, Enric Corona, Gerard Pons-Moll, and Francesc Moreno-Noguer.
\newblock D-nerf: Neural radiance fields for dynamic scenes.
\newblock In {\em CVPR}, 2021.

\bibitem{raj2020anr}
Amit Raj, Julian Tanke, James Hays, Minh Vo, Carsten Stoll, and Christoph
  Lassner.
\newblock Anr: Articulated neural rendering for virtual avatars.
\newblock In {\em CVPR}, 2021.

\bibitem{romero2017embodied}
Javier Romero, Dimitrios Tzionas, and Michael~J Black.
\newblock Embodied hands: Modeling and capturing hands and bodies together.
\newblock {\em ACM ToG}, 2017.

\bibitem{saito2019pifu}
Shunsuke Saito, Zeng Huang, Ryota Natsume, Shigeo Morishima, Angjoo Kanazawa,
  and Hao Li.
\newblock Pifu: Pixel-aligned implicit function for high-resolution clothed
  human digitization.
\newblock In {\em ICCV}, 2019.

\bibitem{saito2020pifuhd}
Shunsuke Saito, Tomas Simon, Jason Saragih, and Hanbyul Joo.
\newblock Pifuhd: Multi-level pixel-aligned implicit function for
  high-resolution 3d human digitization.
\newblock In {\em CVPR}, 2020.

\bibitem{schonberger2016structure}
Johannes~L Schonberger and Jan-Michael Frahm.
\newblock Structure-from-motion revisited.
\newblock In {\em CVPR}, 2016.

\bibitem{shysheya2019textured}
Aliaksandra Shysheya, Egor Zakharov, Kara-Ali Aliev, Renat Bashirov, Egor
  Burkov, Karim Iskakov, Aleksei Ivakhnenko, Yury Malkov, Igor Pasechnik,
  Dmitry Ulyanov, et~al.
\newblock Textured neural avatars.
\newblock In {\em CVPR}, 2019.

\bibitem{sitzmann2019scene}
Vincent Sitzmann, Michael Zollh{\"o}fer, and Gordon Wetzstein.
\newblock Scene representation networks: Continuous 3d-structure-aware neural
  scene representations.
\newblock In {\em NeurIPS}, 2019.

\bibitem{su2020robustfusion}
Zhuo Su, Lan Xu, Zerong Zheng, Tao Yu, Yebin Liu, et~al.
\newblock Robustfusion: Human volumetric capture with data-driven visual cues
  using a rgbd camera.
\newblock In {\em ECCV}, 2020.

\bibitem{thies2019deferred}
Justus Thies, Michael Zollh{\"o}fer, and Matthias Nie{\ss}ner.
\newblock Deferred neural rendering: Image synthesis using neural textures.
\newblock {\em ACM TOG}, 2019.

\bibitem{weng2019photo}
Chung-Yi Weng, Brian Curless, and Ira Kemelmacher-Shlizerman.
\newblock Photo wake-up: 3d character animation from a single photo.
\newblock In {\em CVPR}, 2019.

\bibitem{wu2020multi}
Minye Wu, Yuehao Wang, Qiang Hu, and Jingyi Yu.
\newblock Multi-view neural human rendering.
\newblock In {\em CVPR}, 2020.

\bibitem{xu2020ghum}
Hongyi Xu, Eduard~Gabriel Bazavan, Andrei Zanfir, William~T Freeman, Rahul
  Sukthankar, and Cristian Sminchisescu.
\newblock Ghum \& ghuml: Generative 3d human shape and articulated pose models.
\newblock In {\em CVPR}, 2020.

\bibitem{yang2021s3}
Ze Yang, Shenlong Wang, Sivabalan Manivasagam, Zeng Huang, Wei-Chiu Ma, Xinchen
  Yan, Ersin Yumer, and Raquel Urtasun.
\newblock S3: Neural shape, skeleton, and skinning fields for 3d human
  modeling.
\newblock In {\em CVPR}, 2021.

\bibitem{yoon2021pose}
Jae~Shin Yoon, Lingjie Liu, Vladislav Golyanik, Kripasindhu Sarkar, Hyun~Soo
  Park, and Christian Theobalt.
\newblock Pose-guided human animation from a single image in the wild.
\newblock In {\em CVPR}, 2021.

\bibitem{yu2021plenoctrees}
Alex Yu, Ruilong Li, Matthew Tancik, Hao Li, Ren Ng, and Angjoo Kanazawa.
\newblock {PlenOctrees} for real-time rendering of neural radiance fields.
\newblock In {\em ICCV}, 2021.

\bibitem{zheng2019deephuman}
Zerong Zheng, Tao Yu, Yixuan Wei, Qionghai Dai, and Yebin Liu.
\newblock Deephuman: 3d human reconstruction from a single image.
\newblock In {\em ICCV}, 2019.

\end{thebibliography}
}

\newpage

\setcounter{section}{0}
\section*{Supplementary Material}

In the supplementary material, we provide the derivation of transformation matrices, network architectures, details of training and test data, and 3D reconstruction results.

\section{Derivation of transformation matrices}

We represent the human skeleton as $(\mathbf{J},\boldsymbol{\theta})$, where $\mathbf{J}\in \mathbb{R}^{K\times3}$ denotes the joint locations of $K$ joints and $\boldsymbol{\theta} \in \mathbb{R}^{3(K+1)\times1}=[\boldsymbol{\omega}_0^T,...,\boldsymbol{\omega}_K^T]$ denotes the $(K+1)$ relative rotation of body part with respect to its parent part in a kinematic tree using the axis-angle representation. Then, the transformation matrix of part $k$ from canonical pose $\boldsymbol{\theta}_c$ to target pose $\boldsymbol{\theta}_t$ can be represented as 

\begin{equation}
	\label{eq:motion_basis_f}
	G_k = A_k(\mathbf{J},\boldsymbol{\theta}_t)A_k(\mathbf{J},\boldsymbol{\theta}_c)^{-1},
\end{equation}

\begin{equation}
	\label{eq:motion_basis_g}
	A_k(\mathbf{J},\boldsymbol{\theta}) = \prod_{i \in P(k)}
	\begin{bmatrix}
		R(\boldsymbol{\omega}_i) && \mathbf{j}_i \\
		0 && 1
	\end{bmatrix},
\end{equation}
where $R(\boldsymbol{\omega}_i)\in \mathbb{R}^{3\times3}$ is the converted rotation matrix of $\boldsymbol{\omega}_i$ via the Rodrigues formula, $\mathbf{j}_i$ is the $i$-th joint center, and $P(k)$ is the ordered set of parent joints of joint $k$. In practice, we adopt the SMPL skeleton \cite{loper2015smpl}, which has $K=24$ parts, but this idea applies to other human skeletons \cite{cao2018openpose, ionescu2013human3}.

\section{Network architectures}

\begin{figure}[t]
\centering
\includegraphics[width=1\linewidth]{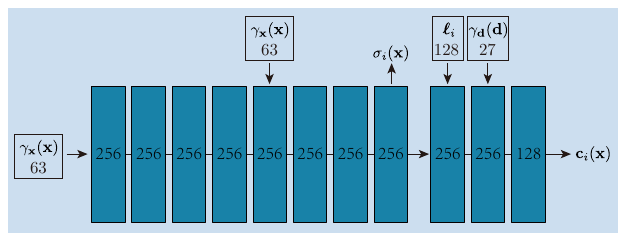}
\caption{\textbf{Network architecture of the density and color fields.} The network is almost the same as the original NeRF, except that we introduce a per-frame latent code $\boldsymbol{\ell}_i$ to encode the state of human appearance in frame $i$. The number in each block means the dimension of the input.}
\label{fig:network}
\end{figure}

\begin{figure}[t]
\centering
\includegraphics[width=1\linewidth]{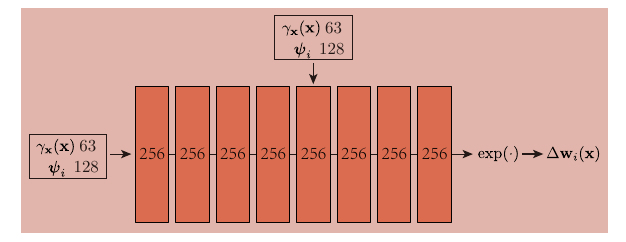}
\caption{\textbf{Network architecture of the neural blend weight filed.} The network takes the positional encoding of the location $\gamma_\mathbf{x}(\mathbf{x})$ along with a per-frame latent code $\boldsymbol{\psi}_i$ and outputs the residual blend weight $\Delta \mathbf{w}_i(\mathbf{x})$ using exponential map. The network consists of 8 linear layers with ReLU activations and includes a skip connection on the fifth layer, which is similar to the density prediction module of the original NeRF. The number in each block means the dimension of the input.}
\label{fig:network2}
\end{figure}

We present architectures of NeRF network and neural blend weight field network in Figures \ref{fig:network} and \ref{fig:network2}, respectively.

\section{Training and test data}

We show the detailed frame numbers for training and test of each subject in Table \ref{table:train_test}. Since the video length of each subject is different, we choose the appropriate number of frames (150$\sim$300) to train the model and take remaining video frames for test.

\begin{table}[!htbp]
\begin{center}
\tablestyle{4pt}{1.05}
\begin{tabular}{x{25}|x{15}|x{15}|x{15}|x{15}|x{15}|x{15}|x{15}}
& S1 & S5 & S6 & S7 & S8 & S9 & S11 \\[.1em]
\shline
training & 150 & 250 & 150 & 300 & 250 & 260 & 200 \\
test & 49 & 127 & 83 & 200 & 87 & 133 & 82 \\
\end{tabular}
\end{center}
\vspace{-1em}
\caption{\textbf{Frame numbers for training and test of each subject of the H36M dataset.}}
\vspace{-1em}
\label{table:train_test}
\end{table}

\section{3D reconstruction}

\begin{figure*}[t]
\centering
\includegraphics[width=1\linewidth]{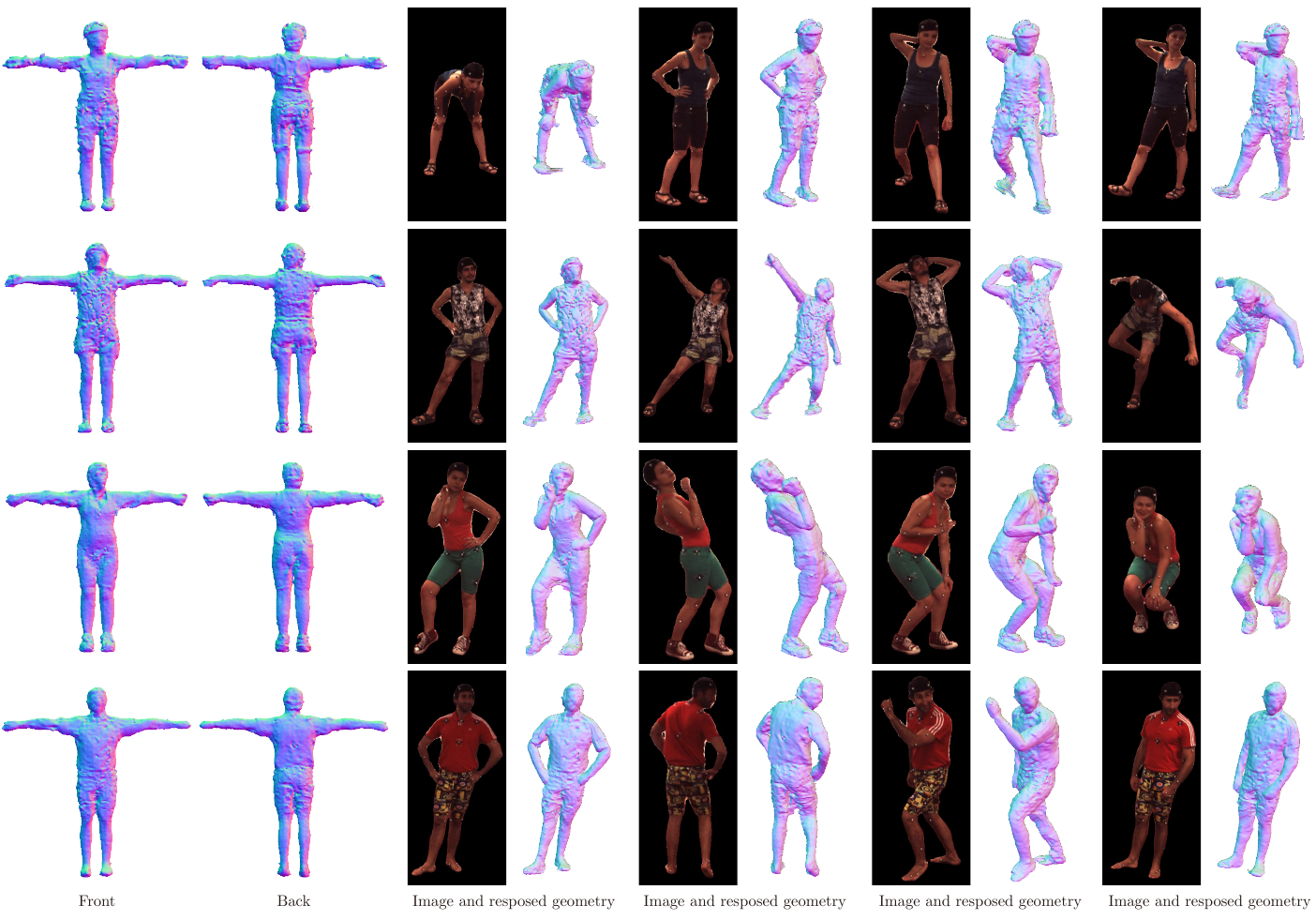}
\caption{\textbf{Reconstructed geometries and reposed geometries.} The first two columns show the reconstructed geometries in the canonical space, which can be animated according to input human poses.}
\label{fig:shape_generation}
\end{figure*}

Figure \ref{fig:shape_generation} presents the reconstruction results in the canonical space in the first two columns. As described in the paper, we can use the learned blend weight field to animate the reconstructed geometry, which is also shown in Figure \ref{fig:shape_generation}. We find that the original reconstruction tend to be rough, which may be caused by the inaccurate segmentation results. To solve this problem, we additionally apply Gaussian smoothing to the reconstructed geometry. We present more results in the supplementary video.

\end{document}